%% file: main.tex
\documentclass[10pt,twocolumn,letterpaper]{article}

\usepackage{cvpr}              

\input{preamble}

\usepackage[sort&compress]{natbib}

\definecolor{cvprblue}{rgb}{0.21,0.49,0.74}
\usepackage[pagebackref,breaklinks,colorlinks,citecolor=cvprblue]{hyperref}

\title{LTGC: Long-tail Recognition via Leveraging LLMs-driven Generated Content}

\author{Qihao Zhao$^{1,2,}$\thanks{Equal contribution.}, Yalun Dai$^{3,*}$, Hao Li$^{4}$, Wei Hu$^{1}$, Fan Zhang$^{1}$\thanks{ The Corresponding author is with the College of Information Science and Technology and the Interdisciplinary Research Center for Artificial Intelligence, Beijing University of Chemical Technology, China}, Jun Liu$^{2}$\\
$^{1}$Beijing University of Chemical Technology, China\\
$^{2}$Singapore University of Technology and Design, Singapore\\
$^{3}$Nanyang Technological University, Singapore\\
$^{4}$Northwestern Polytechnical University, China\\
}

\begin{document}
\maketitle
\input{sec/0_abstract}    
\input{sec/1_intro}
{
    \small
    \bibliographystyle{ieeenat_fullname}
    \bibliography{main}
}
\end{document}

%% file: preamble.tex
\usepackage[dvipsnames]{xcolor}

\usepackage{multirow}
\usepackage{pifont}



%% file: sec/0_abstract.tex
\begin{abstract}

Long-tail recognition is challenging because it requires the model to learn good representations from tail categories and address imbalances across all categories. In this paper, we propose a novel generative and fine-tuning framework, LTGC, to handle long-tail recognition via leveraging generated content. Firstly, inspired by the rich implicit knowledge in large-scale models (e.g., large language models, LLMs), LTGC leverages the power of these models to parse and reason over the original tail data to produce diverse tail-class content. We then propose several novel designs for LTGC to ensure the quality of the generated data and to efficiently fine-tune the model using both the generated and original data. The visualization demonstrates the effectiveness of the generation module in LTGC, which produces accurate and diverse tail data. Additionally, the experimental results demonstrate that our LTGC outperforms existing state-of-the-art methods on popular long-tailed benchmarks. \href{https://ltgccode.github.io}{Project Link.}
\end{abstract}

%% file: sec/1_intro.tex
\section{Introduction}
\label{sec:intro}

In the real world, data often exhibits a long-tailed distribution, posing significant challenges for computer vision recognition \cite{zhang2023deep, longtailwang2017metalearning}. These challenges include (1) Class Imbalance: within the dataset, some classes (termed "head" classes) are abundantly represented, while others (termed "tail" classes) have few samples. This imbalanced distribution during training may cause the model to focus more on the head classes, neglecting the tail classes \cite{wang2020longRIDE}. (2) Tail Data Scarcity: Data scarcity refers to tail classes having an extremely limited number of samples, which lack diversity and are insufficient to effectively train a model \cite{li2021metasaug, chen2022imagine}. It prevents a model's ability to learn the feature invariant, which is necessary to recognize these categories correctly \cite{tang2022invariant}.

\begin{figure}[t]
  \centering
    \includegraphics[width=0.95\linewidth]{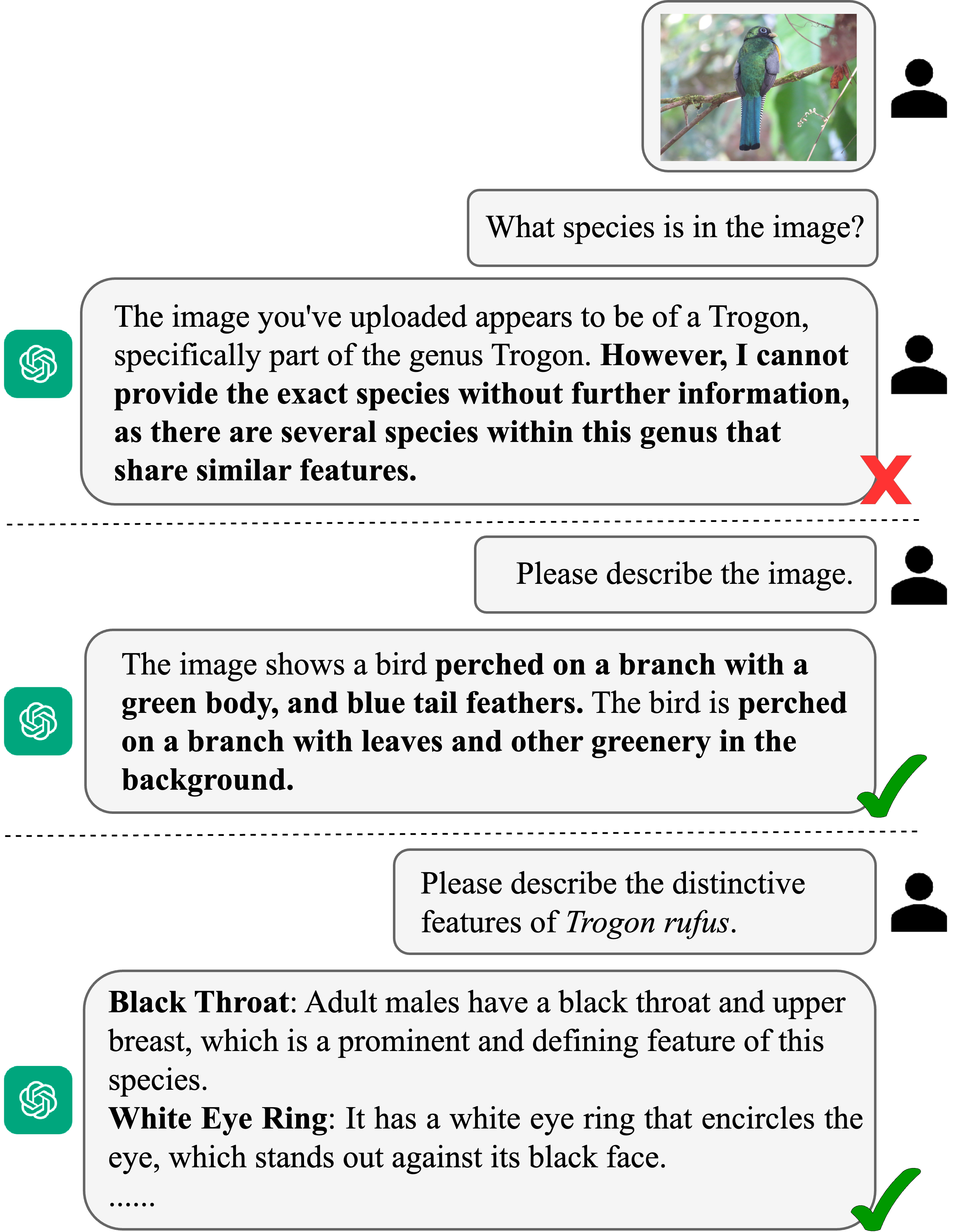}
    \caption{\textbf{Top}: Here, our LMMs use ChatGPT. For the \textit{Trogon Rufus} category, when we asked ChatGPT, "What species is in the picture?" we did not get the expected answer due to the complexity of the question. \textbf{Middle and Down}: In contrast, when we asked some easy questions for the ChatGPT, "Please describe the image." or "Please describe the distinctive features of \textit{Trogon Rufus}." It could accurately answer these questions.}
    \label{fig:fig1}
\vspace{-10px}
\end{figure} 

To address these challenges, numerous approaches have emerged within the field of long-tailed recognition, such as re-sampling \cite{oversamplebuda2018systematic, undersamplejapkowicz2002class}, loss re-weighting \cite{LTkhan2017cost,focalloss,reweightcui2019class,reweightcao2019learning,reweightxie2019intriguing, aimar2023balanced, reweightmenon2020long,liu2019large,9577321}, ensemble learning \cite{xiang2020learning, wang2020longRIDE, zhao2023mdcs}, decoupling \cite{LTkang2019decoupling}, and contrastive learning \cite{wang2021contrastive,zhu2022balanced,cui2021parametric,li2022targeted}. Their primary objective is to balance the decision boundaries and parameter weights of the model to improve long-tail recognition. However, these methods encounter bottlenecks due to the scarcity and limited diversity features of tail-class data \cite{li2021metasaug}. To obtain more diverse representations for tail classes, two types of methods appear: Some methods increase tail-class diversity through data augmentation \cite{chou2020remix, li2021metasaug, zhang2017mixup, zhao2022mixpro}. Others increase tail-class diversity by transferring features from related classes \cite{chen2022imagine} or large pre-trained models (e.g., CLIP \cite{radford2021clip}). These transfer learning methods \cite{dong2022lpt, RAC, tian2022vlltr} based on CLIP have recently shown great potential in boosting long-tail recognition. The LPT \cite{dong2022lpt} fine-tunes pre-trained models to adapt to target datasets. The VL-LTR \cite{tian2022vlltr} leverages text features from CLIP to augment the learning of image features. However, these methods find it hard to obtain the correct and desired knowledge for tail categories. For example, in the real-world data iNauralist 2018 \cite{DBLP:journals/corr/HornASSAPB17}, the category \textit{Aquilegia Pubescens} is characterized only by white or pale yellow colors and spurred shapes. If we transfer similar category features or blindly augment semantic information, such as red color semantics, this category will become more confusing.

Recently, Large Language Models (\textbf{LLMs}) (e.g., ChatGPT) and Large Multimodal Models (\textbf{LMMs}) (e.g., GPT-4V (ChatGPT with Vision \cite{yang2023gpt4v}) and Minigpt4 \cite{zhu2023minigpt, chen2023minigptv2}) due to their wealth of implicit knowledge, have been leveraged for a variety of downstream tasks, such as robot task plans \cite{10161317}, open-set object recognition \cite{qu2023lmc}, automated robot learning \cite{wang2023robogen} and various visual reasoning tasks \cite{yang2023gpt4v, zang2023contextual, zhu2023llafs}.
However, due to the bias of imbalanced training data, LMMs perform poorly on some complex tasks \cite{cui2023holistic}. We also find this dilemma in long-tail recognition as shown in Fig. \ref{fig:fig1}. When we asked LMMs about an image in the category of \textit{Trogon rufus}: "What species is in the image? ", we did not get the expected answer. In contrast, when we ask some easy questions about this image: "Please describe the image." or "Please describe the distinctive features of \textit{Trogon Rufus}", we get the desired answer. This suggests that although large models underperform in long-tail recognition (e.g., it is difficult to align complex image features with labels.), they still contain desired implicit knowledge 
(e.g., correctly describing images and providing the knowledge of species). Inspired by this, we aim to leverage the rich implicit knowledge of several large models to address the challenges of long-tailed data scarcity and perform long-tailed recognition well. Nevertheless, effectively leveraging large models to encounter these challenges is not trivial: (1) The implicit nature of knowledge within large models makes it difficult to extract the desired knowledge that facilitates long-tail recognition tasks.  
(2) Due to the domain gap between the generated data and the original data \cite{trabucco2023effective}, effectively using these hybrid data remains an open issue.

\begin{figure*}[t]
  \centering
    \includegraphics[width=0.9\linewidth]{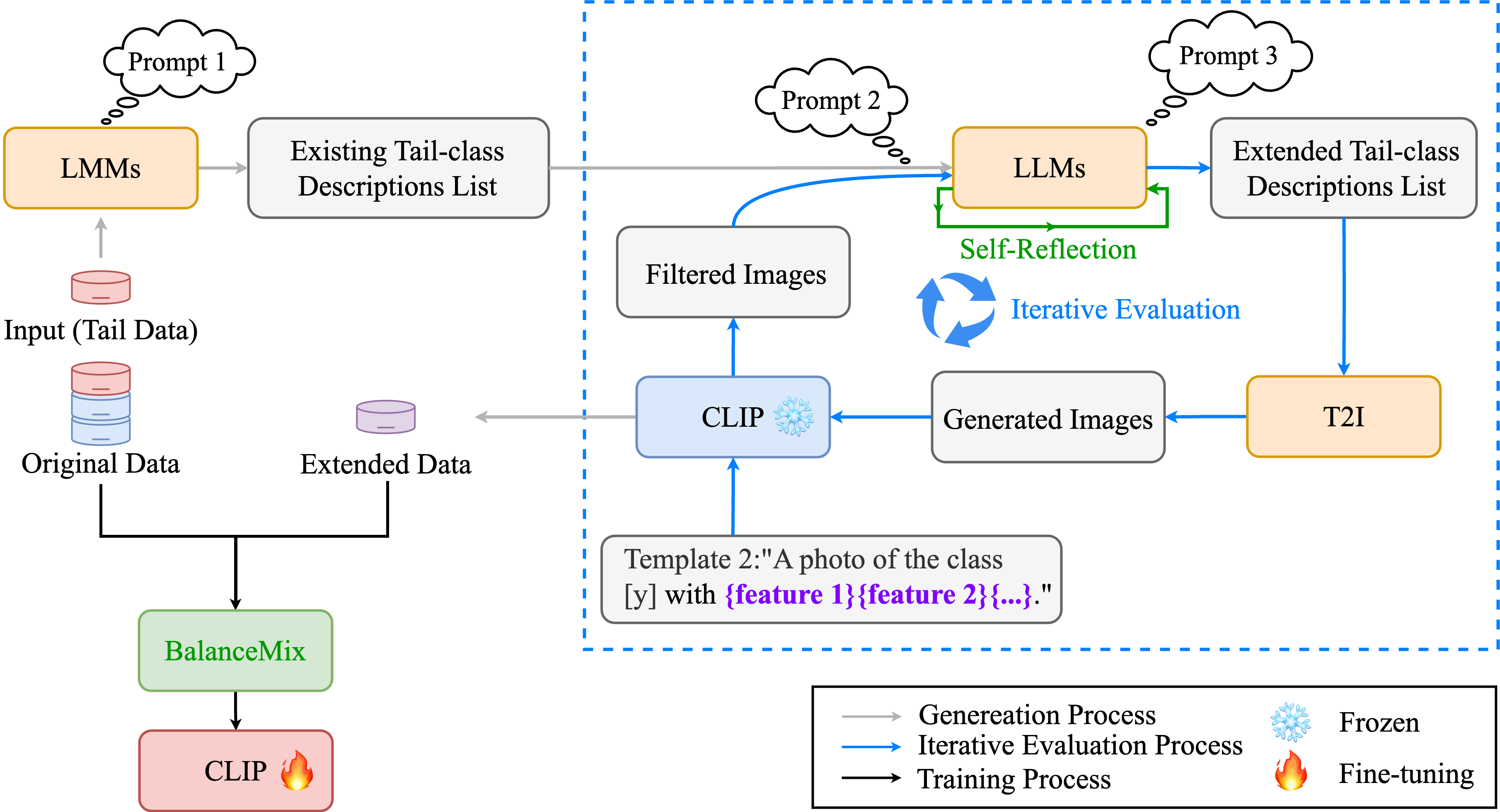}
    \caption{\textbf{Overall framework of LTGC}. LTGC first employs LMMs to analyze the existing tail data to obtain the existing tail-class descriptions list. Then it inputs the list into LLMs to analyze the absent features of the tail classes and employs the T2I model to generate diverse images. Moreover, our designed self-reflection and iterative evaluation modules ensure the diversity and quality of the tail data. Finally, LTGC employs the BalanceMix module to fine-tune the CLIP's visual encoder with the extended and original data.} 
    \label{fig:framework}
\vspace{-10px}
\end{figure*}

To this end, we propose a novel \textbf{L}ong-\textbf{T}ail recognition framework via \textbf{G}enerated \textbf{C}ontent, denoted as \textbf{LTGC}, which is illustrated in Fig. \ref{fig:framework}. Our LTGC aims to leverage the capabilities of large models for generating explicitly diverse content \cite{foo2023aigenerated}  tailored to the long-tail classes and incorporate novel designs to enhance long-tail recognition. Firstly, motivated by the wealth of implicit knowledge in large models and the fact that texts are more controllable, we employ them to expand the tail classes. Specifically, to produce more \textit{diverse and controllable} tail data, we follow the rule of analyzing existing features before generating absent features. To achieve this, LTGC utilizes LMMs to analyze images already in the tail classes to obtain the existing tail-class descriptions list. Then LTGC inputs the existing descriptions list to LLMs to obtain the desired extended tail-class descriptions list for images that are absent from the existing tail classes. Secondly, LTGC utilizes the text-to-image (T2I) model (e.g., DALL-E \cite{ramesh2021zero}) to transform these textual descriptions into images. Moreover, inspired by the benefits of chain-of-thought \cite{wei2022chain}, we propose a self-reflection and iterative evaluation module for this process to ensure the diversity and quality of the generated content. Finally, inspired by the advantages of Mixup \cite{zhang2017mixup} in merging different image domains, we propose the \textbf{BalanceMix} module to address domain shifts of generated images for the fine-tuning process. 

In summary, the contributions of our work are as follows:
\begin{enumerate}
    \item In the \textbf{first} time, we propose a novel framework via generated content, LTGC, which leverages the power of large models to address long-tail recognition challenges.
    \item We design a series of novel modules to tackle the tail-class image scarcity problem and design the BalanceMix module to efficiently fine-tune the model using the generated data and the original data.
    \item Experimental results demonstrate that our LTGC outperforms existing state-of-the-art methods on popular long-tail benchmarks. Also, the visualization illustrates the diversity and controllability of our generated tail images.
\end{enumerate}

\section{Related Work}
\textbf{Long-Tail recognition.} 
Many approaches have been proposed in recent years to address the long-tail recognition challenge, such as resampling \cite{oversamplebuda2018systematic,undersamplejapkowicz2002class}, loss rebalancing \cite{LTkhan2017cost,focalloss,reweightcui2019class,reweightcao2019learning,reweightxie2019intriguing, aimar2023balanced, reweightmenon2020long,liu2019large,9577321} to increase the model's focus on the tail categories, ensemble learning \cite{wang2020longRIDE, zhang2021test, zhao2023mdcs} to improve the ability to recognize tail samples by enhancing model performance, contrast learning \cite{yang2020rethinking,kang2020exploring,wang2021contrastive,zhu2022balanced,cui2021parametric,li2022targeted} and calibration learning \cite{chou2020remix,Zhong_2021_CVPR}to improve the model's decision boundaries, but due to the scarcity of data on the tails, the model still doesn't have enough features to learn class representations for long-tail recognition. Recently, data augmentation-based methods \cite{li2021metasaug, chou2020remix, chen2022imagine} proposed to handle scarcity-tailed data by augmenting from related features. However, these methods find it difficult to obtain the correct knowledge for tail classes. Other approaches \cite{dong2022lpt, tian2022vlltr} have achieved good results by transfer learning \cite{torrey2010transfer} the feature of the visual-language pre-training model (e.g., CLIP \cite{radford2021clip}) for long-tailed recognition.  

Different from the previous methods, in this paper, from a novel perspective, we explore a generative and fine-tuning framework to handle long-tail recognition by leveraging the rich implicit knowledge of large models. Specifically, we design several novel modules to leverage the correct knowledge for generating tail data and perform efficient fine-tuning with the generated data to boost long-tail recognition.

\noindent\textbf{Large Models.}
Recently, several large models have emerged, including large language models (e.g., ChatGPT \cite{yang2023gpt4v}), large multi-modal models (e.g., GPT-4 (ision) \cite{yang2023gpt4v}, MiniGPT \cite{zhu2023minigpt}) and large generated models (e.g., DALL-E \cite{ramesh2021zero}). These expansive models are repositories of extensive knowledge \cite{foo2023aigenerated} and have undergone scrutiny in diverse applications, such as robot task plans \cite{10161317}, open-set object recognition \cite{qu2023lmc}, contextual object detection \cite{zang2023contextual}, video generation \cite{khachatryan2023text2video}, and few-shot segmentation \cite{zhu2023llafs}. In this work, we leverage the rich implicit knowledge of these large models to handle the long-tail recognition challenge.

\section{Method}
To perform long-tail recognition well, the challenge is making the model learn diverse representations from tail categories and address imbalances across all categories.
In this paper, we propose a novel long-tail recognition framework, LTGC, to handle these challenges. As depicted in Fig. \ref{fig:framework}, LTGC employ implicit knowledge from various off-the-shelf large-scale models to generate and iteratively assess the quality and diversity of tail classes (Sec. \ref{subsecgeneration}). Furthermore, LTGC proposes the BalanceMix module to facilitating the fine-tuning process with the generated and original datasets (Sec. \ref{subcecbalancemix}). These modules are described in detail below.

\subsection{Diverse Tail Images Generation}
\label{subsecgeneration}
To learn diversity representations from tail categories, the previous methods have used data augmentation \cite{li2021metasaug, chou2020remix} or transfer learning \cite{dong2022lpt, tian2022vlltr}. However, these works find it difficult to obtain the correct and desired diversity knowledge for tail categories. Inspired by the common-sense knowledge in the LLMs and the fact that textual descriptions are more controllable \cite{yang2023gpt4v, ramesh2021zero}, LTGC takes advantage of these to control the detail and diversity of the generated tail classes. Firstly, to better generate diverse images and control the image detail, LTGC aims to generate images that are absent in the original tail data and represent these images in textual form. Specifically, LTGC employs LMMs to analyze the original tail data and obtain the \textbf{existing tail-class descriptions list}. Then LTGC leverages the common-sense knowledge of LLMs to \textbf{obtain extended tail-class descriptions} based on the existing tail-class descriptions list. Finally, as images are more suitable as training data for long-tail recognition tasks, we utilize the text-to-image module to generate diverse images based on these tail-class descriptions.

\subsubsection{Obtaining Existing Tail-class Descriptions List}
\label{subsecobtain}
In order to ensure the diversity of tail-class images, we first analyze the feature information of the original tail-class images before generating new ones. It guarantees that the content of the generated images is distinct from that of the existing tail-class images. Moreover, due to textual descriptions being more controllable \cite{ramesh2021zero}, we utilize the textual descriptions to control the detail and diversity of the generated images. To achieve this, our LTGC employs Language Model Multimodals (LMMs), such as GPT-4V (Vision) \cite{yang2023gpt4v}, to analyze the feature information and extract textual descriptions of existing tail classes. During this process, the textual responses from LMMs could be varied and sometimes redundant, which may impede the generation of the desired images. Therefore, we employ textual templates to constrain the responses of LMMs, aiming to unify the textual description formats. Inspired by an image of an object that can be fully described or generated by its class and a list of features \cite{besserve2019counterfactuals, wang2021self}, we design the textual template to include the given class and its features. Furthermore, the introduction of variations in the scenes plays a crucial role in enhancing the model's ability to generalize \cite{arjovsky2019invariant}, a problem that's particularly acute in long-tailed datasets \cite{tang2022invariant}. Therefore, the textual information of scenes is also important for image generation.

To this end, we design the LMMs' response \textbf{Template 1} for a given class $y$ as follows:
{\fontfamily{qcr}\selectfont
"A photo of the class [$y$], \{\textcolor[RGB]{0,0,255}{with distinctive features}\}\{\textcolor[RGB]{0,153,0} {in specific scenes}\}."}, which include the given class, its distinctive features, and specific scenes or environments. With the response template, for a given $M_y$ number class $y$, we sequentially feed these tail-class images into LMMs along with an instructing \textbf{[Prompt 1]} to analyze the features of these images:
{\fontfamily{qcr}\selectfont
"Please use the \textbf{Template 1} to briefly describe the image of the class $[y]$."}
This process is illustrated in Fig. \ref {fig:minigpt}, where $y$ is the given class label. As shown, by formulating the description using the template, LMMs would automatically replace the class, features, and background in the response. After performing this process for all images of each class, we compile a list of text descriptions corresponding to each class. Then we intend to extend the tail-class descriptions list with the existing tail-class descriptions list to generate the images lacking in the tail classes.

\subsubsection{Obtaining Extended Tail-class Descriptions List}
\label{subsectextdescription}
In this section, we aim to analyze features missing from tail classes and enrich the descriptions of these classes based on the existing tail-class descriptions list. To accomplish this, inspired by the rich common-sense knowledge of Large Language Models (LLMs), such as ChatGPT \cite{yang2023gpt4v}, we extend the tail-class descriptions list through a two-step process: 1) Inputting the existing descriptions list into LLMs, and 2) Designing the \textbf{Prompt 2} to guide LLMs in generating the desired descriptions for images that are absent in the given tail class $y$:
{\fontfamily{qcr}\selectfont
"Besides these descriptions mentioned above, please use the \textbf{Template 1} to list other possible \{\textcolor[RGB]{0,0,255}{distinctive features}\} and \{\textcolor[RGB]{0,153,0}{specific scenes}\} for the class $[y]$,"} which is illustrated in Fig. \ref {fig:chatgpt}. For each class, we repeat the above two-step process, and then we obtain the tail-class descriptions list for all tail classes.

In addition, in order for these generated descriptions to better complement each tail class, we encourage LLMs to generate descriptions of sufficient number and diversity. To achieve this, we introduce a \textbf{self-reflection} module in this process, aiming to guide LMMs in rethinking if there are any features or scenes that are missed or repeated. It includes two key designs: a number-checking module and a repetition-checking module.
(1) The number-checking module aims to guide LMMs in rethinking whether there are other missing features or scenarios. To achieve this, after posing the initial \textbf{[Prompt 2]} for class $y$, we update the extended descriptions list and re-ask LLMs the \textbf{[Prompt 2]} question, incorporating the newly acquired list. For each class $y$, this iterative process of number-checking continues until a maximum number $K_y$ of the tail class is achieved, where $K_y$ = $M_y$+$N_y$ and $N_y$ is the number of generated descriptions for class $y$.
(2) The repetition-checking module aims to guide LMMs in rethinking if there are other features or scenes that are repeated at the end of the number-checking iteration. Specifically, we input the extended descriptions list and the following \textbf{[Prompt 3]} of each class $y$ for LLMs' repetition checking:
{\fontfamily{qcr}\selectfont"Please exclude any repetitive \{\textcolor[RGB]{0,0,255}{distinctive features}\} and \{\textcolor[RGB]{0,153,0}{specific scenes}\} for class $[y]$ in this descriptions list."} After that, LLM will filter out the ones with repeated features based on this prompt and return a new list of descriptions. Through the implementation of this two-step process and LLMs' self-reflection module, we have obtained a reduction in repeated descriptions and an increase in the diversity of the tail-class descriptions for each class.

\subsubsection{Transform Descriptions to Images}
\label{subsecimageextension}
Above, we obtain the extended tail-class descriptions list for each class $y$, denoted as $L_y$. As the images are better adapted to perform visual recognition tasks, in this section, we aim to leverage the image-generative ability of the text-to-image (T2I) method to generate the images from the tail-class descriptions list. In detail, we employ T2I to generate images based on the descriptions list, denoted as:

\begin{figure}[t]
  \centering
 
    \includegraphics[width=0.95\linewidth]{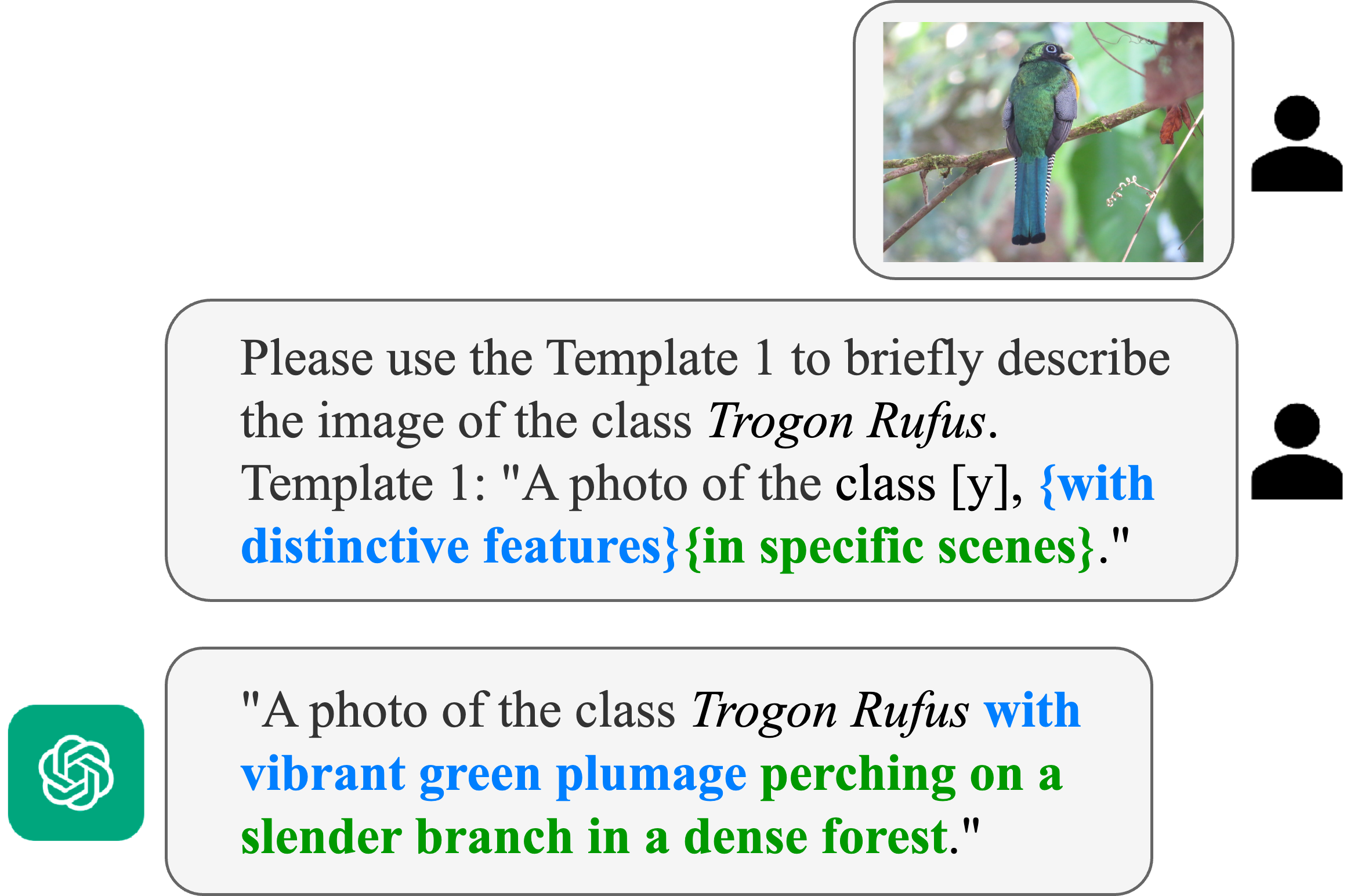}
    \caption{\textbf{Example of the instruction for LMMs.} When both images from tail classes and textual templates are input into LMMs, textual descriptions corresponding to the images can be obtained. By repeatedly performing this operation on the training data, we convert abstract image descriptions into concrete textual descriptions. Finally, we acquire the current textual descriptions list corresponding to each class.}
    \label{fig:minigpt}
\vspace{-10px}
\end{figure} 

\begin{figure}[ht]
  \centering
 
    \includegraphics[width=0.95\linewidth]{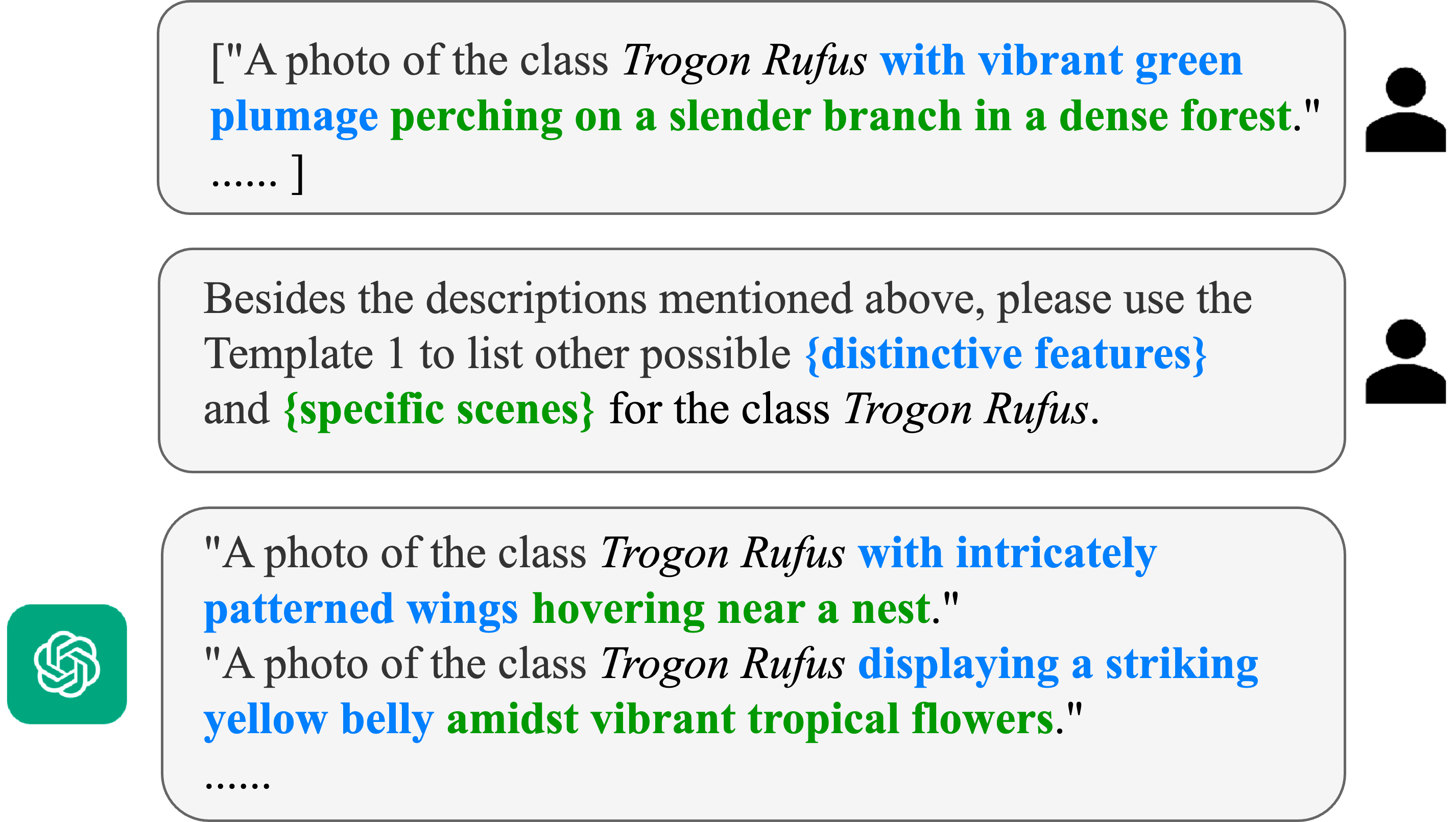}
    \caption{\textbf{Example of the instruction for LLMs.} LTGC inputs the existing textual descriptions list to LLMs, which continually extends it with new distinctive features and scene information. During multiple iterations, LTGC generates a new extended textual descriptions list for each class.}

    \label{fig:chatgpt}
\vspace{-10px}
\end{figure} 

\begin{equation}
i_{n}^{y} = \text{T2I} (d_{n}^{y}), \text{where } n \in \{1, \dots, N\}
\end{equation}
where $n \in N$ denotes the $n$-th generated descriptions for class $y$, and $i_{n}^{y}$ denotes the image generated by T2I model based on $d_{n}^{y}$. However, some generated images may not be of sufficiently high quality to accurately represent their desired classes, as the \textit{lower-quality} images. As shown in Fig. \ref {fig:vis2}, \textit{lower-quality} images will output poorly distinctive features, resulting in more confusion for the class. Therefore, using these generated images directly in training may result in disrupting the model's prediction of the tail classes.

\begin{figure}[ht]
  \centering
    \includegraphics[width=0.98\linewidth]{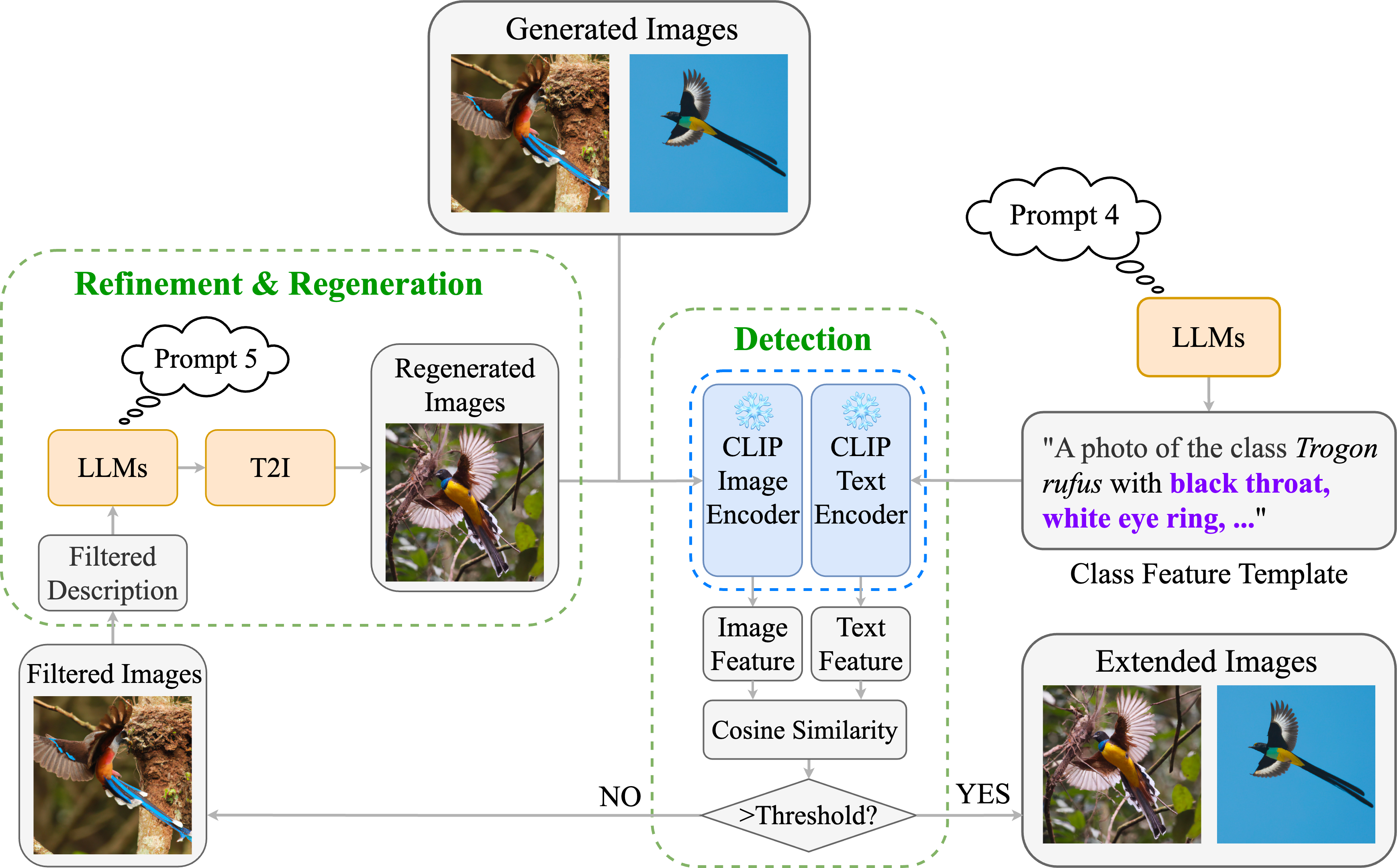}
    \caption{\textbf{Illustration of the proposed iterative evaluation module framework.} This module detects lower-quality images through the similarity score $\mathcal{S}$ computed by images and their corresponding class feature template. Then the textual descriptions corresponding to \textit{lower-quality} images are re-input into LLMs for refinement. Finally, the refined textural descriptions are fed into the T2I model for regeneration.}
    \label{fig:filter}
\vspace{-10px}
\end{figure} 

To address the challenge of \textit{lower-quality} images that fail to represent their desired classes accurately, we aim to automatically detect these images, refine their textual descriptions, and regenerate them accordingly. To achieve this, drawing inspiration from the idea that humans improve their understanding through peer feedback, we investigate the possibility of a large model also enhancing its output by integrating feedback from another fundamental model. Therefore, we propose a \textbf{iterative evaluation} module, as illustrated in Fig. \ref{fig:filter}. This module utilizes the fundamental model, CLIP, to serve as the 'feedback giver' selected for its proficiency in understanding images and effectively connecting them with textual contexts \cite{radford2021clip}. It provides feedback about the \textit{lower-quality} images to LLMs, which assists in refining their corresponding descriptions.

Furthermore, to facilitate CLIP in grasping the represented features in images, a straightforward way is to check the degree of match between the image and its class template: "A photo of a [$\text{y}$]." However, due to the fact that CLIP's text encoder may contain less feature knowledge of the tail classes (e.g., as presented on the iNaturalist 2018 dataset with poor zero-shot recognition performance \cite{dong2022lpt}), we aim to design a \textbf{new} textual feature template for each class, highlighting its unique features. Then CLIP employs it to determine the correspondence between the template and the image. Drawing inspiration from the chain-of-thought \cite{wei2022chain} concept, which suggests that LMMs perform better when provided with additional clues, we further guide LMMs in summarizing the most distinctive features of each class. Specifically, after obtaining the extended tail-class descriptions list, we obtain the \textbf{class feature template} $C_y$ that contains the most distinctive features by the following \textbf{[Prompt 4]} and \textbf{Template 2} for the given class $y$:
\textbf{[Prompt 4]}: {\fontfamily{qcr}\selectfont"Please use \textbf{Template 2} to summarize the most distinctive features of class $[y]$\}."}
\textbf{Template 2}: {\fontfamily{qcr}\selectfont"A photo of the class $[y]$ with \{\textcolor{purple}{feature 1}\}\{\textcolor{purple}{feature 2}\}\{\textcolor{purple}{...}\}."} This process could guide the LMMs to rethink and summarize the most distinctive features and produce the class feature template $C_y$ for each class $y$. Furthermore, since we focus on checking whether the feature information of an image could represent its desired class, we do not include scene information in the class feature template. Then we introduce the details of the iterative evaluation module.
\label{sec:cyclic-assessing}

As shown in Fig. \ref{fig:filter}, our iterative evaluation module conducts the following three steps iteratively: (1) Detection: To identify \textit{lower-quality} images, we first employ utilize the strong capability of aligning images with text of CLIP to match the feature of generated images $i_{n}^y$ and its class feature template $C_y$ by cosine similarity metric as follows:

\begin{equation}
\mathcal{S} = \text{Encoder}_{\text{vis}}(i_{n}^y) \cdot \text{Encoder}_{\text{text}}(C_y), 
\end{equation}
where $\text{Encoder}_{\text{vis}}$ denotes the CLIP’s visual encoder, and $\text{Encoder}_{\text{text}}$ denotes the text encoder of the CLIP . The image $i_{n}^y$ is detected as \textit{lower quality} and filtered out if its similarity score $\mathcal{S}$ is below a threshold $\mu$. 
\textbf{2) Refinement.} Next, also inspired by chain-of-thought \cite{wei2022chain}, we propose the following prompt for refine its descriptions. Specifically, if $i_{n}^y$ is identified as a \textit{lower quality} image, 
we prompt LLMs to refine its corresponding description $d_{n}^y$ to more accurately represent the intended class $y$, drawing on feedback provided by CLIP, i.e., \textbf{[Prompt 5]}:
 
{\fontfamily{qcr}\selectfont
"This description $d_{n}^y$ doesn’t seem to be representative of the class $[y]$. Could you refine it to enhance the distinctive features of class $[y]$?"}
\textbf{3) Re-generation.} Finally, the image $i_{n}^y$ is regenerated by the T2I model according to the improved textual description.

By iteratively applying these three stages, we employ CLIP in each cycle to furnish precise feedback to the LLMs, thereby steering the refinement of descriptions to better align with our desired direction. These refined descriptions ensure the production of images that more accurately embody the characteristics of each class.

\subsection{BalanceMix}
\label{subcecbalancemix}
With generated tailed images, the final problem is how to efficiently use these generated data and original data to perform long-tailed recognition well. Due to the domain gap between the generated data and the original long-tail data \cite{trabucco2023effective}, we propose a method named BalanceMix to handle this challenge. We first define the original data and generated data as $\mathcal{D}_{o}$ and $\mathcal{D}_{g}$. Then BalanceMix balance-sample \cite{oversamplebuda2018systematic} an image $x_i$ from $\mathcal{D}_{o}$ and sample an image $x_j$ from  $\mathcal{D}_{g}$. Meanwhile, it mixes the images $x_i$ and $x_j$ and their corresponding labels, denoted as:
\begin{equation}\label{equCELoss}
  \widetilde{x} = \lambda \odot x_i + (\textbf{1}-\lambda) \odot x_j,
  \end{equation}
  \begin{equation}
  \widetilde{y} = \lambda \odot y_i + (\textbf{1}-\lambda) \odot y_j,
\end{equation}
where $\lambda$ in Beta (0,1) distribution. Finally, we fine-tune the CLIP's vision encoder with LORA \cite{hu2021lora} on all mixed data pairs ($\widetilde{x}, \widetilde{y}$) for efficient long-tail recognition.

\section{Experiments}
We present the experimental results on three widely used datasets in long-tailed recognition, including ImageNet-LT \cite{liu2019large}, Places-LT \cite{liu2019large}, and iNaturalist 2018 \cite{DBLP:journals/corr/HornASSAPB17}. Moreover, we undertake ablation studies specifically on the ImageNet-LT and iNaturalist 2018 datasets to gain deeper insights into the performance of our method. The experimental results of the comparison methods are taken from their original paper, and our results are averaged over three experiments. 

\subsection{Implementation details.}
\textbf{Evaluation Setup.} In all experiments, we evaluate and report top-1 accuracy on their corresponding test set. We also report accuracy on three splits of the classes: Many-shot (more than 100 images), Medium-shot (20 to 100 images), and Few-shot (less than 20 images) \cite{LTkang2019decoupling}.

\noindent\textbf{Method Implementation.} 
In our LTGC, we incorporate diverse and specialized knowledge from the off-the-shelf large models. Specifically, for LMM, we use the GPT-4V (ision) \cite{yang2023gpt4v} version of ChatGPT. For LLM, we use the GPT-4 version of ChatGPT. For T2I, we use DAll-E \cite{ramesh2021zero}. For the pre-trained CLIP \cite{radford2021clip}, we use ViT-B/32 \cite{vit} for its visual encoder and the transformer architecture described in \cite{radford2019language} for its text encoder. 
In LLM's self-reflection module, we set the maximum number $K_y$ to 100, 300, and 800 for iNaturalist 2018, ImageNet-LT, and Place-LT, respectively.
In the iterative evaluation module, the threshold $\mu$ is set at 0.8 for ImageNet-LT and Place-LT, and at 0.6 for iNaturalist.
In Appendix, we provide a detailed ablation and discussion on the choice of LMM and LLM models, as well as parameters used in the self-reflection process.

\subsection{Comparisons with SOTA on Benchmarks}
In this section, we compare our proposed LTGC model with state-of-the-art (SOTA) methods on three benchmarks, including Imagenet-LT \cite{liu2019large}, Places-LT \cite{liu2019large}, and iNaturalist 2018 \cite{DBLP:journals/corr/HornASSAPB17}. To ensure a fair comparison, we primarily focus on methods based on CLIP, as they also leverage knowledge pretrained on large-scale datasets. These baselines include CLIP Zero-Shot \cite{radford2021clip, tian2022vlltr} and CLIP Finetune \cite{tian2022vlltr}, as well as CLIP-based long-tail recognition approaches such as VL-LTR \cite{tian2022vlltr}, LPT \cite{dong2022lpt}, and RAC \cite{RAC}.
Additionally, we report comparisons with traditional methods (without CLIP) on the challenging and fine-grained, large-scale long-tail dataset, iNaturalist 2018. These results demonstrate the effectiveness of our method in different scenarios.

\noindent\textbf{Results on Imagenet-LT}. In Tab. \ref{tab:imagenetandplace}, we observe that our LTGC models are superior to other CLIP-based LT methods. For example, the overall accuracy of our method reaches 80.6\%, which outperforms the existing SOTA method, VT-LTR, \cite{tian2022vlltr} by 3.4\%. 
Moreover, the overall accuracy of our method marginally surpasses the results on the full ImageNet (i.e., 80\% \cite{tian2022vlltr}).

\noindent\textbf{Results on Places-LT}. Tab. \ref{tab:imagenetandplace} shows that compared to other CLIP variant methods, LTGC achieves 54.1\% and 52.1\% in terms of overall accuracy and few-shot accuracy, respectively surpassing the LPT \cite{dong2022lpt} by 4.0\% and 5.2\%.
Even compared with VL-LTR \cite{tian2022vlltr} and RAC \cite{RAC}, which have extra data in training and testing, our LTGC achieves remarkable results.

\noindent\textbf{Results on iNaturalist 2018}.
Finally, we explore LTGC on a large-scale and fine-grained dataset, iNaturalist 2018. Tab. \ref{tab:inaturalist} presents the quantitative results. LTGC leverages the rich knowledge of LLMs and significantly outperforms traditional deep-learning approaches for long-tail recognition. In addition, LTGC attains an overall accuracy of 82.5\% and a few-shot accuracy of 82.6\%, outperforming all existing SOTA methods based on CLIP.
In particular, LTGC also surpasses the retrieval augmented method, RAC \cite{RAC}, by 2.3\%. 

\subsection{Compare with different methods of LMMs.}
\label{exp:lmms}
In the experiments, our LTGC employs GPT-4V (ision) to describe a given image to obtain a text-based feature description. A straightforward way for long-tail recognition is by querying LMMs for the category of the given image.
To evaluate the ability of these methods to recognize the long tail images, we constructed the following baselines of LMMs for comparison: MiniGPT4 \cite{zhu2023minigpt}, MiniGPT4-v2 \cite{chen2023minigptv2}, LENS \cite{berrios2023lens}, and GPT-4V (ision) \cite{yang2023gpt4v}. To perform a fair evaluation, we provide the label list [Class 1, Class 2, ..., Class Y] for each dataset, where $Y$ is the number of classes. 

The results presented in Tab. \ref{tab:llms}, show that our method significantly outperforms the baselines that directly query the LMMs. 
More implementation details are discussed in Appendix.

\begin{table}[ht]
\footnotesize
\centering
\caption{Comparison with SOTA methods on ImageNet-LT and Places-LT.}
\label{tab:imagenetandplace}
\begin{tabular}{c|cc|cc}
\hline
                        Dataset       & \multicolumn{2}{c|}{ImageNet-LT} & \multicolumn{2}{c}{Places-LT}          \\ \hline
                               &Few   &All     &Few &All                           \\ \hline \hline
                        CLIP Zero \cite{tian2022vlltr}         &58.6 &59.8	&40.1 &38.0         \\
                        CLIP Finetune \cite{tian2022vlltr}     &34.5 &60.5	&22.7 &39.7	 	    \\  \hline
                        VL-LTR \cite{tian2022vlltr}            &59.3 &77.2	&42.0 &50.1 	    \\
                        RAC \cite{RAC}                         &-	&-	    &41.8 &47.2	        \\
                        LPT \cite{dong2022lpt}                 &-	&-	    &46.9 &50.1 	    \\ \hline
                        LTGC(Ours)                             &\textbf{70.5} &\textbf{80.6}	&\textbf{52.1} &\textbf{54.1}         \\
                        \hline

\end{tabular}
\vspace{-10px}
\end{table} 

\begin{table}[ht]
\footnotesize
\centering
\caption{Comparison with SOTA methods on iNaturalist 2018.}
\label{tab:inaturalist}
\begin{tabular}{c|cccc}
\hline
                        Method       &Many   &Medium     &Few &All                           \\ \hline \hline
                        Softmax                          &74.7	&66.3	&60.0  &64.7         \\
                        LADE \cite{hong2021disentangling}     &64.4	&47.7	&34.3  &52.3         \\
                        RIDE \cite{wang2020longRIDE}         &71.5	&70.0	&71.6  &71.8         \\
                        PaCo \cite{cui2021parametric}         &69.5	&73.4	&73.0  &73.0         \\
                        MDCS \cite{zhao2023mdcs}         &76.5	&75.5	&75.2  &75.6         \\\hline
                        CLIP Zero \cite{tian2022vlltr}         &6.1	&3.3	&2.9  &3.4         \\
                        CLIP Finetune \cite{tian2022vlltr}  &76.6	&74.1	&70.2 &72.6 	   \\  \hline
                        VL-LTR \cite{tian2022vlltr}       &-	&-	&-  &76.8	   \\ 
                        RAC \cite{RAC}                   &75.9	&80.5	&81.0 &80.2	   \\
                        LPT \cite{dong2022lpt}          &-	&-	&79.3   &76.1	   \\ \hline
                        LTGC(Ours)                      &\textbf{77.5}	&\textbf{83.9}	&\textbf{82.6} &\textbf{82.5}      \\
                        \hline

\end{tabular}
\vspace{-10px}
\end{table} 

\begin{table}[ht]
\footnotesize
\centering
\caption{Comparison with different LMMs' methods on ImageNet-LT and iNaturalist2018.}
\label{tab:llms}
\begin{tabular}{c|c|c}
\hline
                         Method                 &ImageNet-LT   &iNaturalist 2018           \\ \hline
                         LENS \cite{berrios2023lens}                   &69.5     	&17.4	   \\
                         MiniGPT4 \cite{zhu2023minigpt}               &60.4		  &20.9	   \\
                         MiniGPT4-v2 \cite{chen2023minigptv2}            &68.5     	&27.1	   \\ 
                         GPT-4                   &72.1     	&64.3	   \\ \hline
                         Ours                   &\textbf{80.6}		&\textbf{82.5}	   \\  \hline

\end{tabular}
\vspace{-15px}
\end{table} 

\subsection{Analysis and Ablation Study}

\begin{figure*}[ht]
  \centering
 
    \includegraphics[width=0.90\linewidth]{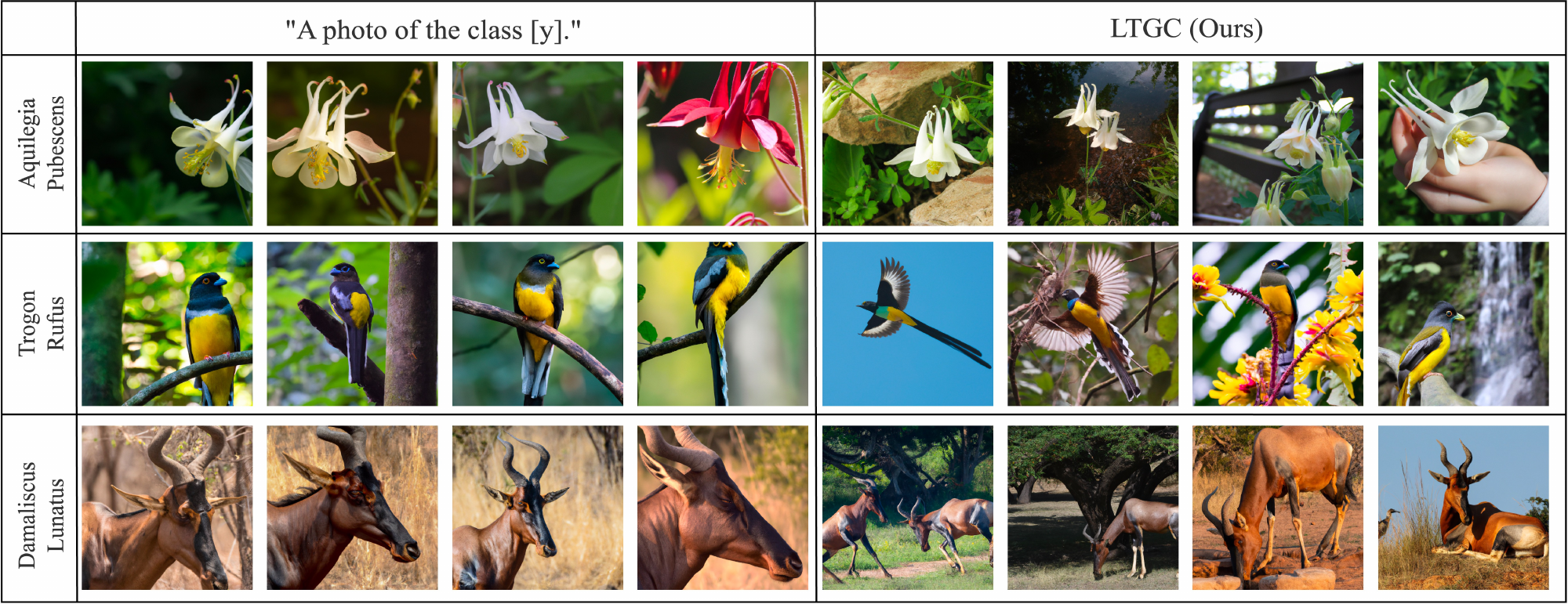}
    \caption{\textbf{The visualization of generated images: The template "A photo of the class [$\text{y}$]” and LTGC.} Each row represents a different class. The four images on the left are generated using the simple template "A photo of the class [$\text{y}$]," which results in images with uniform poses and plain backgrounds. The four images on the right are from the proposed LTGC and demonstrate the diversity of classes.}  
    \label{fig:vis1}
\vspace{-10px}
\end{figure*} 

\textbf{Effectiveness of the iterative evaluation module.}
To guarantee the accurate representation of the desired classes by the images produced via T2I, we have integrated an iterative evaluation module within our architecture for the progressive refinement of images.
To assess the effectiveness of this module, we contrasted it with three distinct image generation strategies:
1) w/o iterative evaluation: the images are fed directly into our framework's training process without any preliminary detection or refinement.
2) Detection and exclusion: the CLIP model evaluates the generated images, selectively forwarding only the ones that align closely with the intended class criteria to the training phase. Images that fail to meet the detection threshold are excluded, bypassing the refinement step entirely.
As illustrate in Tab. \ref{tab:ablation1}, the performance of the two variants is worse than our method. This suggests that our proposed iterative evaluation module incorporating filtering and refinement of the design is more effective.
 
\begin{table}[ht]
\footnotesize
\centering
\caption{Evaluation on the effectiveness of the iterative evaluation.}
\label{tab:ablation1}
\begin{tabular}{c|c|c}
\hline
                         Method                 &ImageNet-LT   &iNaturalist 2018            \\ \hline
                         w/o iterative evaluation      &55.8		&64.9	   \\
                         Detection and exclusion     &71.5     	&77.4	  \\ \hline
                         Ours                      &80.6		&82.5	  \\  \hline
\end{tabular}
\vspace{-1px}
\end{table} 

\vspace{-10px}
\begin{figure}[ht]
  \centering
    \includegraphics[width=0.99\linewidth]{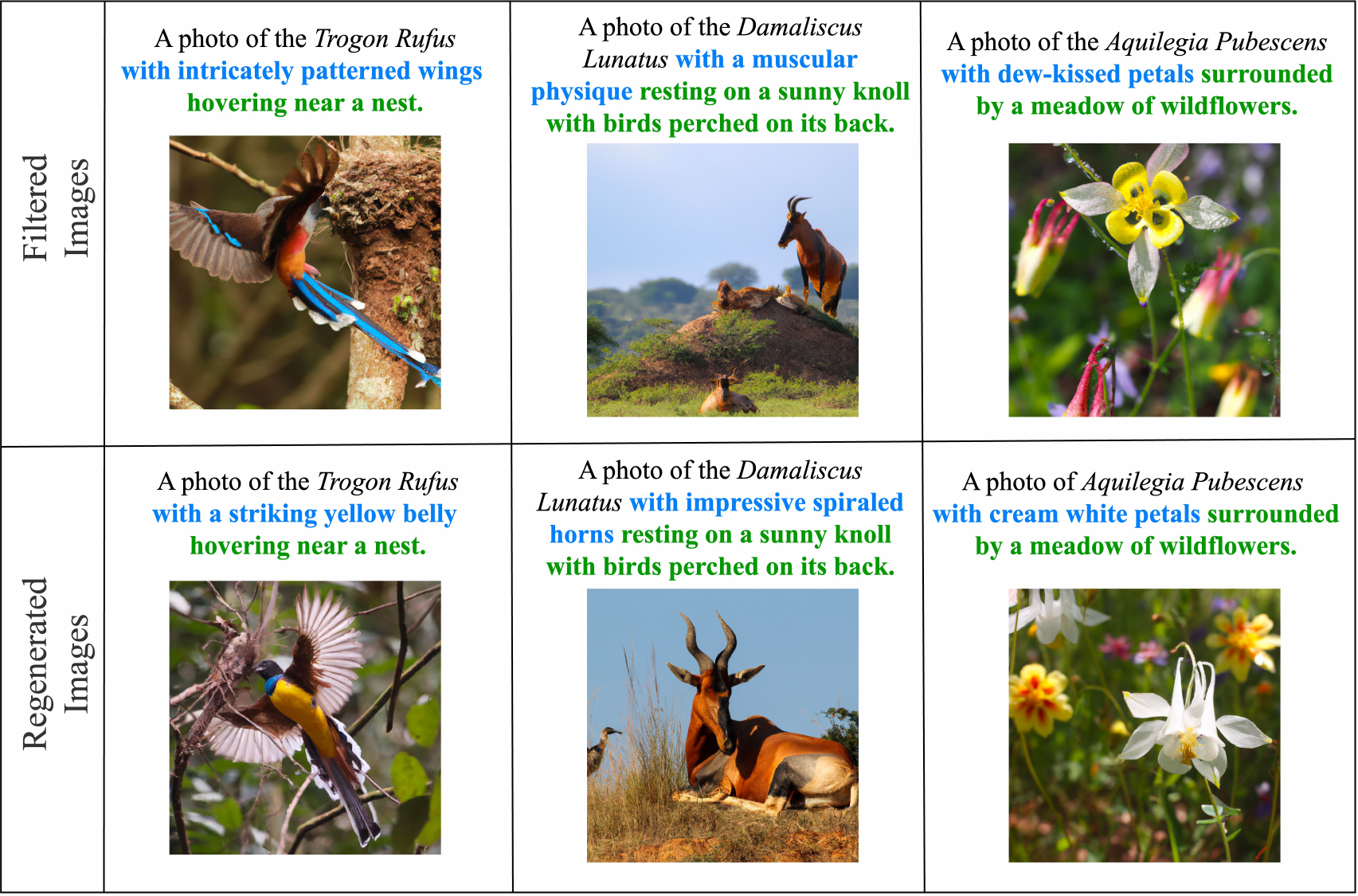}
    \caption{\textbf{The visualization of the images generated before and after passing the iterative evaluation module.} The top row displays images that were filtered out, while the bottom row shows images regenerated by T2I after refining their corresponding descriptions. More visualizations are in Appendix.} 
    \label{fig:vis2}
\vspace{-15px}
\end{figure} 

\noindent\textbf{Effectiveness of the BalanceMix module.}
To evaluate the effectiveness of the two proposed designs, we conduct experiments on three different variants:
1) w/o BalanceMix: This variant is directly fine-tuned using the generated data and the original training data.
2) Balanced Sample: This variant utilizes generated data to perform balanced sampling \cite{wang2017learning} for training.
3) Mixup: This variant employs generated data with Mixup \cite{zhang2017mixup} without performing balanced sampling.
The results are shown in Tab. \ref{tab:blsmix}, demonstrating that our method significantly outperforms the other variants.
Although our self-reflection and iterative evaluation modules already ensure the diversity and high quality of the generated images, there still exists a domain gap between the generated and original data \cite{trabucco2023effective}.
This domain gap could exacerbate the long-tail effect on the test set (discussed in Appendix).  
A simple balanced sample \cite{wang2017learning} approach fails to solve the class imbalance caused by the mixing of original and generated data, while the Mixup \cite{zhang2017mixup} alone cannot address inter-class imbalances. To this end, our BalanceMix module combines the strengths of both methods, making the generated data well-suited for our framework.

\vspace{-10px}
\begin{table}[ht]
\centering
\footnotesize
\caption{Evaluation on the effectiveness of the BalanceMix.}
\label{tab:blsmix}
\begin{tabular}{c|c|c}
\hline
                         Method                 &ImageNet-LT   &iNaturalist            \\ \hline
                         w/o BalanceMix      &58.3		&69.5	   \\
                         Balanced sample \cite{wang2017learning}    &63.9     	&73.8	  \\ 
                         Mixup \cite{zhang2017mixup}              &73.4     	&75.2	  \\  \hline
                         Ours                &80.6		&82.5	  \\  \hline
\end{tabular}
\vspace{-15px}
\end{table} 

\subsection{Visualization}
\noindent\textbf{Visualization of Generated Images: Template "A photo of the class $[\text{y}]$" vs our LTGC.}
Furthermore, we contrast the images generated by LTGC with a simple prompt "A photo of the class $[\text{y}]$" for the T2I model. As Fig. \ref{fig:vis1} illustrates, LTGC generates more accurate and diverse images compared to those generated by the simple prompt. For example, in the absence of control over the category description, the T2I model generates red features that do not belong to class \textit{Aquilegia Pubescens}. 
In addition, with rich text for image generation, our approach also generates more diverse and accurate images compared to images generated by a simple template. 

\textbf{Visualization on iterative evaluation module.}
We compare images before and after refinement using the iterative evaluation module, and the visualization results are shown in Fig. \ref{fig:vis2}. It shows that images before refinement often possess ambiguous semantic information, and the features of the corresponding classes are not distinct. However, after the refinement process, the quality of the images is substantially enhanced, and the distinctive features of the corresponding classes become more pronounced.

\section{Conclusion}
This paper introduces a novel generative and fine-tuning framework, named LTGC, to address the challenge of long-tail recognition. LTGC leverages the abundant implicit knowledge embedded in large-scale models to generate diverse data for tail categories. The framework incorporates innovative designs to ensure the quality of the generated data and to fine-tune the model efficiently using both the generated and original data. The experimental results indicate that LTGC surpasses existing state-of-the-art methods on well-known long-tail benchmarks. In the future, will explore the robustness of large-scale models for application in other areas, such as long-tail semantic segmentation and dissent detection.

\clearpage